\documentclass[letterpaper, 10 pt, conference]{ieeeconf}  

\IEEEoverridecommandlockouts                              

\usepackage{cite}
\usepackage{amsmath,amssymb,amsfonts}
\usepackage{algorithmic}
\usepackage{graphicx}
\usepackage{textcomp}
\usepackage{xcolor}
\usepackage{caption}
\usepackage{makecell}
\usepackage{comment}
\usepackage{subcaption}
\usepackage{paralist}

\title{\LARGE \bf
Designing the mobile robot Kevin for a life science laboratory}

\author{Sarah Kleine-Wechelmann$^{1}$, Kim Bastiaanse$^{2}$, Matthias Freundel$^{1}$ and Christian Becker-Asano$^{2}$
\thanks{}
\thanks{$^{1}$Sarah Kleine-Wechelmann and Matthias Freundel are with the Fraunhofer Institute for 
Manufacturing Engineering and Automation IPA, Stuttgart, Germany
        {\tt\small sarah.kleine-wechelmann@ipa.fraunhofer.de, matthias.freundel@ipa.fraunhofer.de}}%
\thanks{$^{2}$Kim Bastiaanse and Christian Becker-Asano are with the Hochschule der Medien,
        Stuttgart, Germany
        {\tt\small bastiaanse.kim@gmail.com, becker-asano@hdm-stuttgart.de}}%
}

\begin{document}

\maketitle
\thispagestyle{empty}
\pagestyle{empty}

\begin{abstract}

 Laboratories are being increasingly automated. In small laboratories individual processes can be fully automated, but this is usually not economically viable. Nevertheless, individual process steps can be performed by flexible, mobile robots to relieve the laboratory staff.
As a contribution to the requirements in a life science laboratory the mobile, dextrous robot Kevin was designed by the Fraunhofer IPA research institute in Stuttgart, Germany. Kevin is a mobile service robot which is able to fulfill non-value adding activities such as transportation of labware. This paper gives an overview of Kevin's functionalities, its development process, and presents a preliminary study on how its lights and sounds improve user interaction.

\end{abstract}

\section{INTRODUCTION}
\label{introduction}

\begin{figure}[h!]
\centering
\includegraphics[height=210pt]{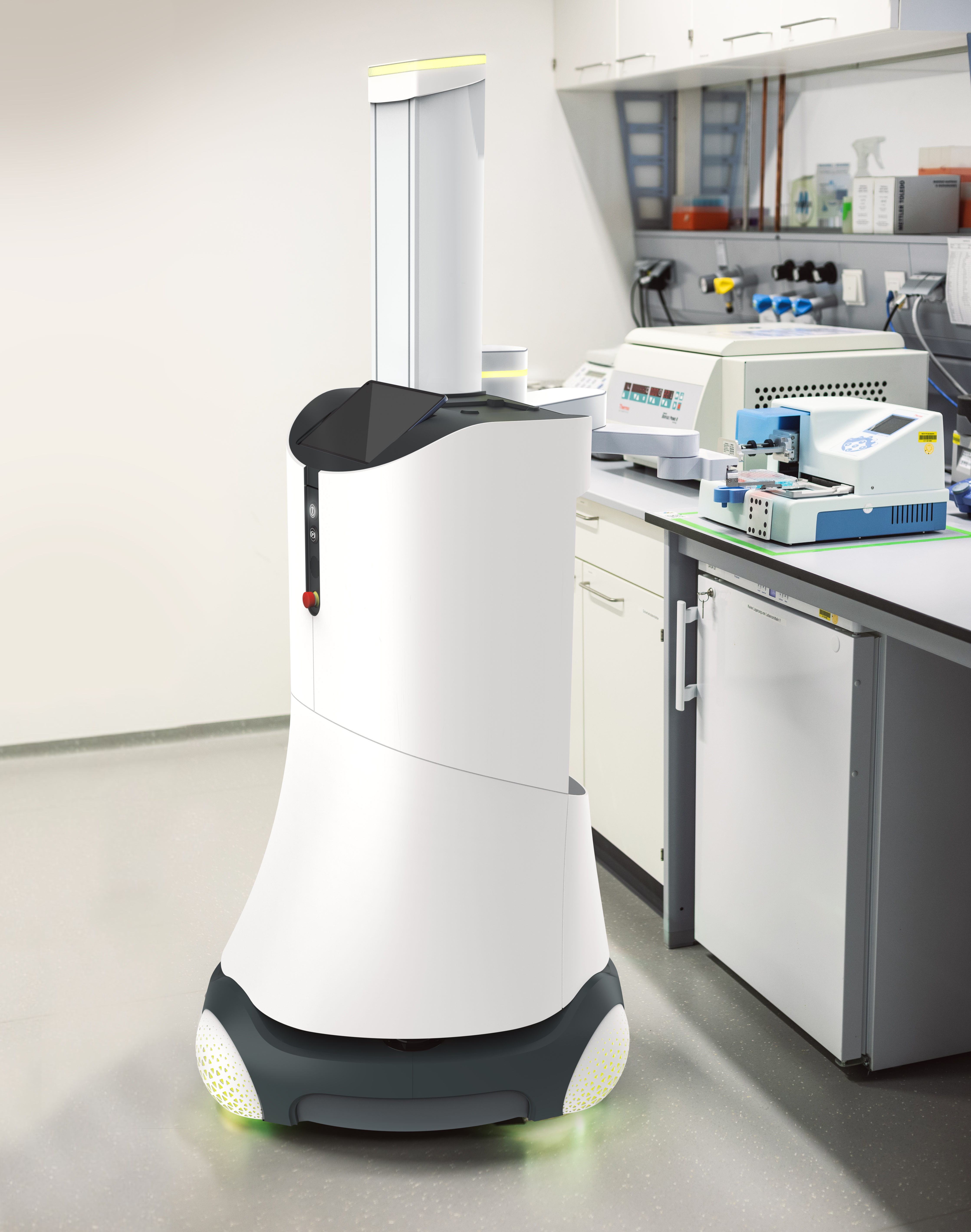}
\caption{Robot Kevin handling a microtiter plate in the laboratory.}
\label{fig:kevinTitle}
\end{figure}

Robots are increasingly present in people's everyday lives, both in professional and private environments. Robots help perform menial and repetitive tasks from the assembly of cars in factories to the cleaning of floors in common households.
In small laboratories, such as those in the life sciences or research and development, individual processes can be fully automated, but usually this is not economically viable.
Nevertheless, individual process steps can be performed by flexible, mobile robots to exonerate the laboratory staff \cite{repetitiveTask_2016}. In effect, the user and the robot share the same workspace, working side by side. For a user, in such a situation, it might not be perceivable which action the robot is currently performing. However, for both the robot’s functioning and the user’s experience, this knowledge is very important so that users can adapt their behavior based on the robot’s specific actions \cite{adaptBehavior_2019}. Therefore, it is important to establish clear methods of communication from the robot to the surrounding users.

This paper describes how the mobile robot Kevin cf.~Figure~\ref{fig:kevinTitle} is designed in order to meet the requirements of a life science laboratory. In addition, this paper presents a preliminary study that was conducted to identify interaction design requirements. Subsequently, the related work is discussed in order to contextualise the design decisions for Kevin.

\section{RELATED WORK}
\label{relatedWork}

 According to the classification provided by Fong et al. \cite{fong2003survey}, a mobile laboratory robot like Kevin can be described as \emph{socially embedded}, because it is supposed to directly interact with humans in a shared environment that it is structurally coupled with. Through its appearance, a robot conveys its affordance to its counterpart: it is important to design this affordance in a purposeful way in order to direct the expectations of the user \cite{norman2013}. Thereby, users should always be able to infer the robot's behavior and function from its morphology \cite{dautenhahn2002design}. This robot morphology can be  designed as humanoid, animoid (zoomorphic), or mechanoid \cite{Groom,Herberg}.


Using the BWIBot familiy of robots, Fernandez et al. \cite{fernandez2018passive} argue that passive demonstrations of light-based robot signals improve the human interpretability of a robot's directional intention in narrow corridors. Their experimental setup features a very technical design of an LED light signal, similar to the work of Baraka and Veloso \cite{baraka2018mobile} on revealing their robot's state through expressing lights. In both cases, however, the interactions with the robots are mainly taking place in corridors of an office environment and not in a biological laboratory as is the case for the robot Kevin.


The use of (non-)verbal utterances and sounds to enhance social human-robot interaction opens up a multitude of opportunities and risks \cite{yilmazyildiz2016review}. The options range from non-vocal sounds and acoustics generated by robots to clearly comprehensible speech expressions. In between these two extremes lies the class of paralinguistic utterances that were employed in Schwenk and Arras' work to flexibly convey their robots' internal states using a real-time synthesizer for sound generation \cite{SchwenkArras2014}. In contrast to our work, they evaluated their results out-of-context and, thus, they might not generalize very well.

According to Cha et al. \cite{nonverbalSignals_2016}, limited expressive capabilities of robots leads to the employment of new signalling methods. The combination of both light and sound signals together with motion for conveying a humanoid robot's emotion has been also evaluated in \cite{haring2011creation}. Using the three-dimensional Pleasure-Arousal-Dominance space of emotional meaning the authors found, that a NAO robot can express the four emotions "anger", "fear", "joy", and "sadness" multimodally. However, the robot was evaluated completely out of context. 

This is problematic because the subjective interpretation of social signals is known to depend on both the observer's cultural background and the situational context \cite{becker2011studying, ishiguro2001robovie}.

\section{CONTEXT OF USE}
\label{laboratory}

Kevin's field of application is in laboratories working in the field of life sciences. There are various tasks and experiments carried out in life science laboratories (e.g. for developing active substances, checking the quality of production and manufacturing products for personalised medicine). Despite the importance of these tasks, laboratory work today is mainly characterised by manual activities and only a few processes are digitalised or automated.
 
The following  laboratory-specific requirements apply for a mobile robot:

\begin{itemize}
    \item \textbf{Active interaction capabilities}: Most laboratories are designed for humans, therefore they and are often narrow, crowded, etc. Sometimes laboratory robots like Kevin will have to operate side by side with a human at the same working station. The robot may also encounter humans in narrow passages. A high-level representation of its surrounding might be necessary in the robot's control software to avoid interaction problems\cite{fong2003survey}.
    \item \textbf{Passive interaction capabilities}: With the robot sharing its environment with the lab staff, it needs to possess interactive capabilities. This primarily involves the transmission of system status information, as the robot is usually controlled by external software systems. Since laboratory personnel usually has no technical training, it is imperative that the status of the robot and its actions can be decoded intuitively. 
    \item \textbf{Problem of noise}: In life science laboratories, the background noise can vary from a low hum to loud noise comparable to that of a factory floor. In addition, even in the case of a low noise environment, e.g.~ with a permanent humming, irregular beeps of the devices have to be dealt with. These environmental conditions complicate the use of sound signals for the laboratory robot Kevin to communicate its status information.
    \item \textbf{Adaption to further circumstances}: Experiments in labs are conducted in many cases around the clock and also on weekends. The robot Kevin has to work at standard working times when there are many people in the laboratory, the sun is shining through the windows and the laboratory lights and devices are switched on. The same applies to the night, when no people are present and the laboratory is nevertheless illuminated brightly with active devices or even without any lighting. This requires sensor technology that enables reliable orientation and navigation despite the various lighting scenarios in a variety of laboratories. However, the robot is not only subject to environmental requirements; cleanability, protection against liquids and of the expensive equipment must also be ensured.\\
    \end{itemize}
    \raggedbottom
    \section{MOTIVATION}
    \label{motivation}
    
    The high variance of tasks and processes requires a high degree of flexibility in the devices and software solutions used and can rarely be met by classic automation. Bespoke technical solutions are costly - often too costly for laboratories working under high financial pressure, hence constantly seeking to maximize the throughput. For these reasons, laboratories try to utilize the existing infrastructure to an ever greater extent. 
    This is even more important in the subclass of biological laboratories. Here, biological cells have to be cared for around the clock. Without automation, this would mean expensive additional shifts for laboratory staff or a forced limitation of performance capacity.

    Activities that do not add value include transporting samples between two workstations, such as an incubator and a microscope, or when nutrient media have to be changed. These repetitive tasks can be taken over by a mobile transport robot that features a mobile base, a manipulator, a storage (so-called "hotel") and a navigation unit. 
    \raggedbottom

\section{TWO STEP DEVELOPMENT PROCESS} 
\label{development}

\subsection{Proof of concept: Kevin 1.0} 
\label{kevin1}

\begin{figure}[h]
\centering
\includegraphics[height=220pt]{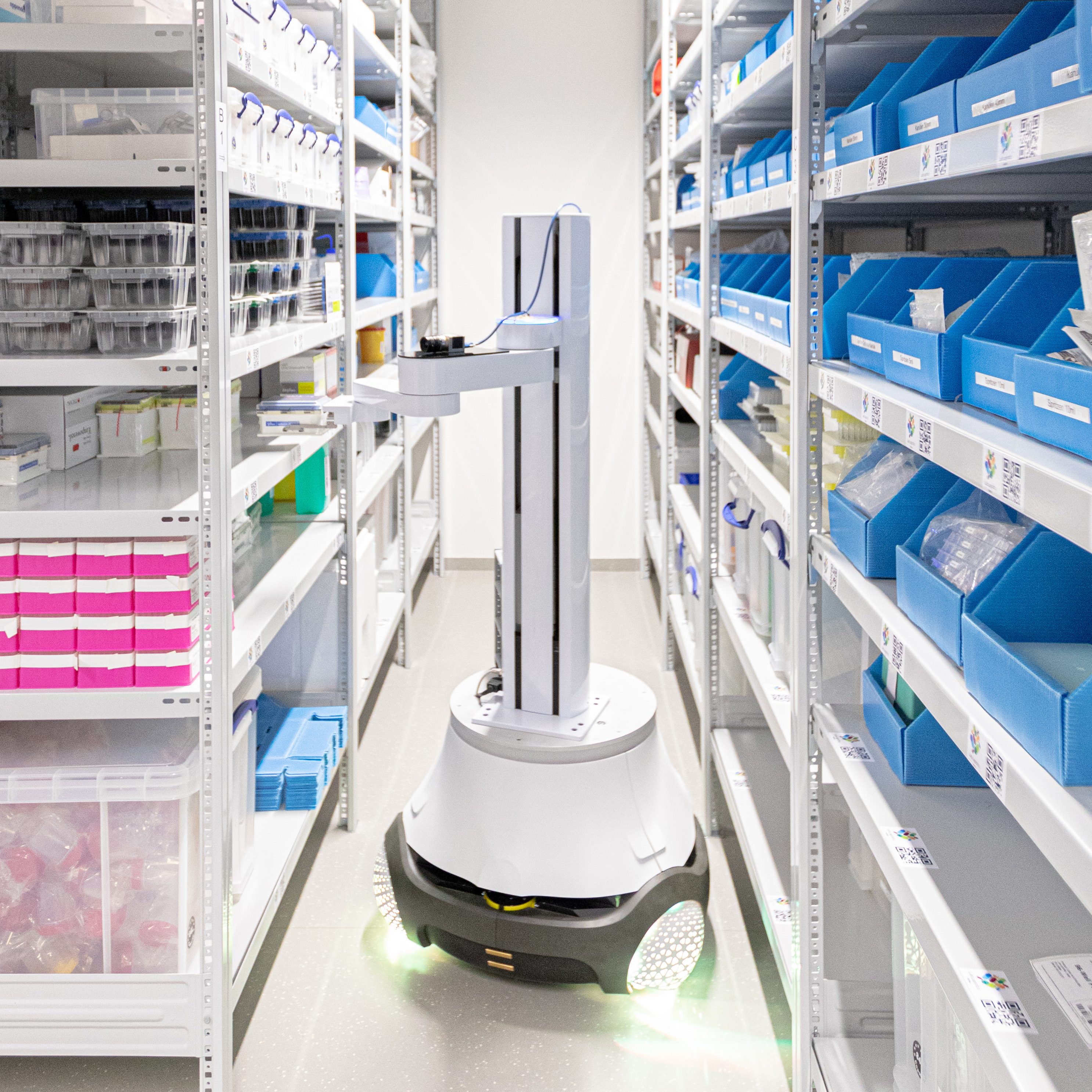}
\caption{Kevin 1.0: Handling labware in a storage.}
\label{fig:kevin}
\end{figure}

We finished designing our first prototype of the laboratory robot Kevin in 2019. With this prototype we aimed to link automation islands, stand-alone equipment and manual workstations day and night. In the laboratory, most systems and devices are designed for microtiter plates with SBS format (127.76 mm × 85.48 mm)\cite{SBSDimensions}. They are used, for example, for cell cultivation or screening of technical bioreactions. The market offers several stationary robotic arms that can handle this format  \cite{Roboticarm2011}. A mobile robot like Kevin, however, is designed to offer far more flexibility.

\subsubsection{Manipulator, mobility, and item transportation}
At the beginning of the robot's development, a highly technology-oriented approach was taken. The well-known four-axis SCARA manipulator of the company Precise was selected for the robot because of its precise positioning capabilities.

To achieve the robot's desired mobility, the base of the Care-o-Bot (service robot for the home environment, cf.~\cite{kittmann2015let}) with omnidirectional drive was used which allows autonomous operation in dynamic and narrow environments. It features a footprint of approximately 70 x 70 cm. Also, it allows Kevin to perform linear movements in all directions, rotation around the center axis point and parallel shift of the track in the moving state. 

\subsubsection{Placement of objects while moving}
For safety reasons, it was decided that the robot should not hold any objects in the gripper while driving. Accordingly, a storage position was necessary. A simple transport position was installed on the robot for this particular purpose.

\subsubsection{Sensors}
In the prototype version, three laser scanners included in the base at ankle height were used to constantly scan the surroundings and enable collision-free navigation. Furthermore, a camera was mounted on the moving part of the arm to allow object manipulation through TAG recognition.

During the technology-oriented design process of this first prototype the following challenges arose:
\begin{enumerate}
    \item Due to the high variance in appearance and the individuality in the control and interaction of robots, various technically correct solutions are possible without the appearance being of the same standard or look. As a result, there is a wide scope for designing the appearance, behavior, performance, and capabilities of robots \cite{dautenhahn_2013}. 
    \item To be accepted by users in the long term and to offer added value, robotic systems must meet the requirements and needs of the users\cite{aRobotCompanion}. For the general acceptance of robots, it is necessary to select the optimal appearance and communication modalities for the corresponding target group and situational context \cite{Blow}.
\end{enumerate}
In advancing from Kevin 1.0 to Kevin 2.0 we addressed these challenges in a systematic fashion.

\subsection{Product: Kevin 2.0} 
After completing the validation of the basic function, i.e., the transport and the manipulation of plates in SBS format in laboratory environments, it was decided to develop a product from the prototype setup Kevin 1.0. The following requirements were gathered for the design into a product:

 \begin{figure}[h]
\centering
\includegraphics[height=140pt]{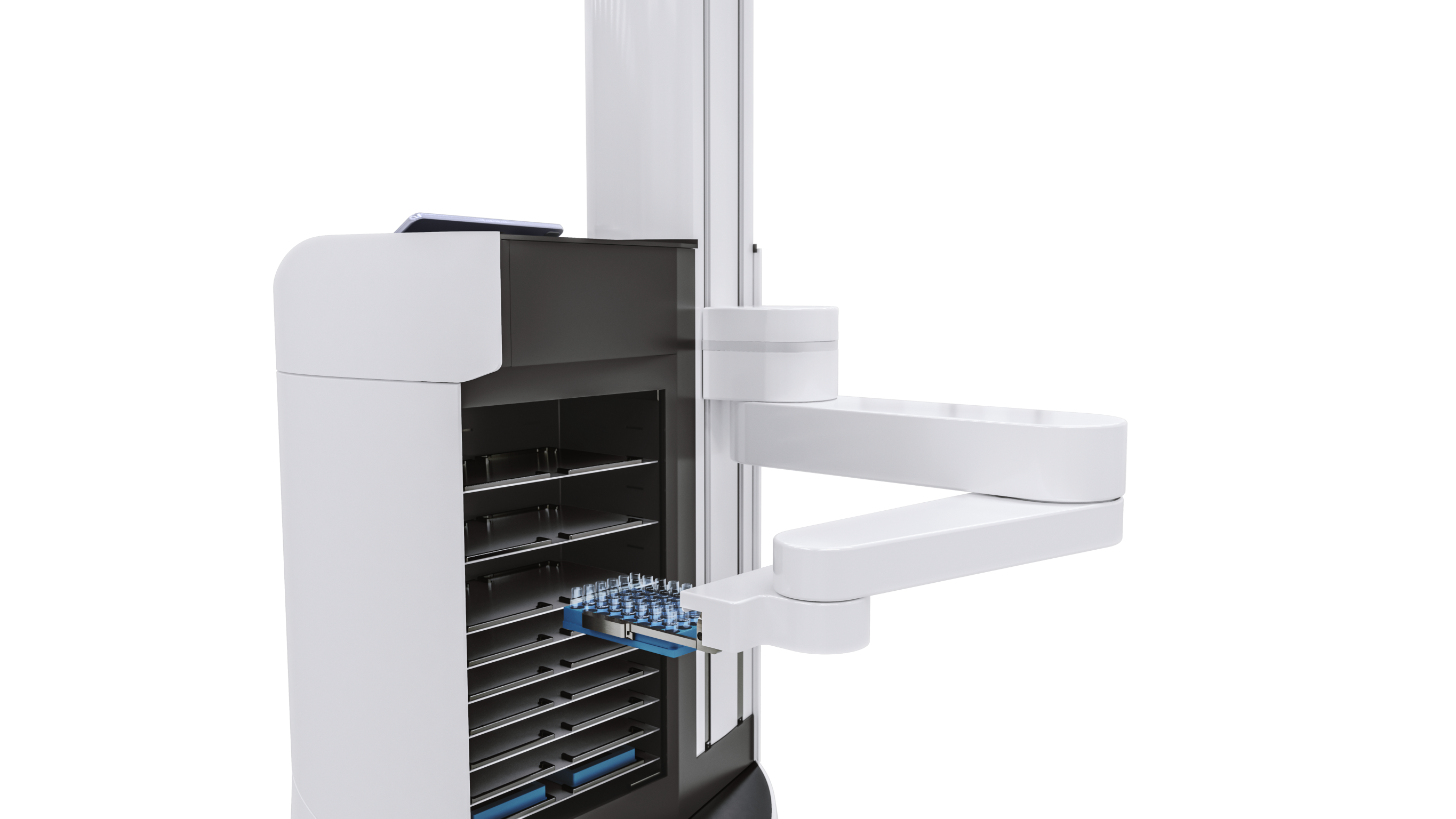}
\caption{Kevin 2.0's Hotel: Kevin uses its Hotel to store a microtiterplate}
\label{fig:hotel}
\end{figure}

\begin{figure}[h]
\centering
\includegraphics[height=150pt]{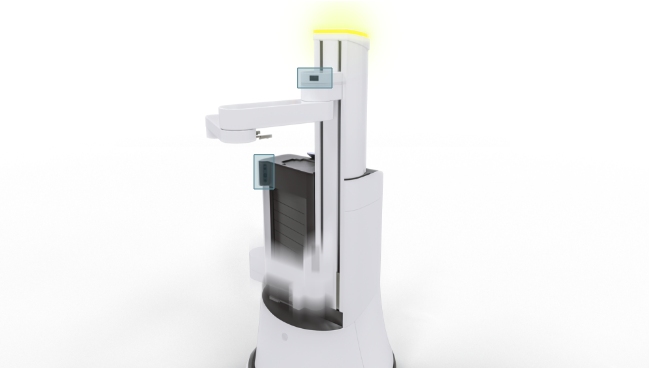}
\caption{Kevin 2.0's Cameras: Kevin has a camera at the shoulder and a camera at the hotel}
\label{fig:cameras}
\end{figure}

\begin{itemize}
    \item The robot must comply with various guidelines such as laboratory equipment guideline DIN EN 61010-1.  
    \item Extension of the transport position: a reconfigurable lockable hotel with higher capacity, cf.~Fig.~\ref{fig:hotel} 
    \item Communication elements for displaying the behavior: lights on different heights visible from all sides, sound output device, removable touch display with teaching interface for configuration and for information about the states e.g. error messages, current task etc. 
    \item Housing to protect the technology e.g. against splash water and designed for the use in sterile environments
    \item Easily accessible operating elements: on/off button, emergency stop switch, brake button
    \item Re-grip station for the robotic arm to reorient plates (landscape or horizontal).  Additionally, the station serves as the primary interaction point for users when manually handing off plates to and from Kevin
    \item Perception: additional camera at the hotel which is used to navigate towards e. g. laboratory devices. With the camera already mounted on the arm, the robot is able to manipulate objects cf.~Fig.~\ref{fig:cameras}
    \item Standardized software interface for linking to a variety of external control solutions
\end{itemize}
 During the development of the second version of the robot Kevin, a holistic approach was chosen, in which the application context and the future user group shaped the entire development process. To be accepted by users in the long term and to offer added value, robotic systems must meet the requirements and needs of the users\cite{mci/Pollmann2019}\cite{aRobotCompanion}. 

\subsubsection{Initial survey}
At the beginning of the development of Kevin 2.0, interviews and online surveys were conducted to include certain parameters by a broad opinion of the potential user group. A total of twenty-nine participants representing the end-users of laboratory robots like Kevin participated in the online survey. In addition, eight people from the life-science industry were interviewed on the same topics as in the online survey. The most important points raised in the initial survey with a total of 37 responses (online survey 29, interviews 8) are presented next. 

The participants were first asked about their prior knowledge of robots in general and specifically about Kevin. They were not shown the function and appearance of version 1.0 in the questioning. When asked how a mobile collaborative robot could support the participants in their daily work, they mentioned the following use cases, among others:
\begin{itemize}
    \item the transport of objects (e.g. transport across zones (leaving sterile area/ clean room), connect warehouse with laboratory, connecting different laboratories with each other)
    \item information provision
    \item logging of activities
\end{itemize}

The consulted potential user group was presented with different scenarios of direct and indirect interaction.
For example, they were asked which communication channels they would use to submit a task to Kevin, or how Kevin should signal that it is approaching the user. Based on these scenarios, the participants were requested to choose the modalities they preferred for the interaction. The majority agreed on the following modalities: external device/ display, light, voice or sound input/ output, movement and positioning and haptic modalities (buttons etc.).

\begin{figure}[ht]
\centering
\includegraphics[width=0.45\textwidth]{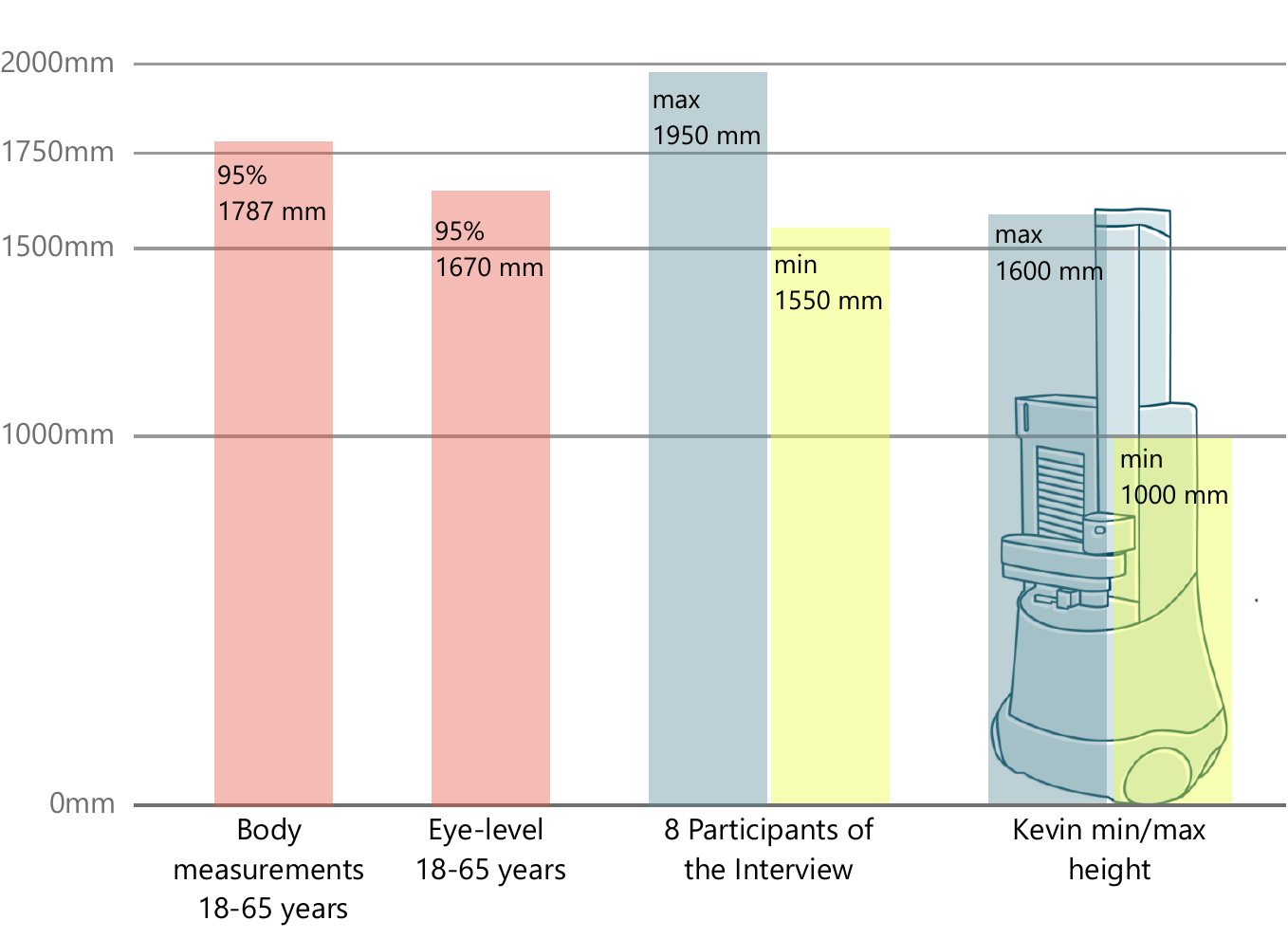}
\caption{Kevin 2.0´s height: Determining the appropriate height for Kevin based on \cite{Lange2013} and the survey of the eight interview participants.}
\label{fig:kevinSize}
\end{figure}

As a result, it was decided to pursue an iconic/ mechanoid design with organic shapes for the robot Kevin. Anthropomorphic modalities and design were not considered appropriate. The color scheme had to follow the bright laboratory environment and clearly show soiling. Furthermore the size of the robot had to be decided which is known to be a crucial aspect of human-robot interaction \cite{Rae2013}. 
Once possible use cases for the Kevin robot had been explained to the participants (see above), they were asked to indicate their preferred size. As a result, the participants suggested that a minimum size of approx. 100 cm and a maximum size of approx. 160 cm would be acceptable. Even with the maximum height, a human interlocutor's eye level is most likely higher than the robots height, considering an average European person, cf.~Figure~\ref{fig:kevinSize}. This is also in line with the design of other interactive, humanoid robots such as Robovie \cite{ishiguro2001robovie}, which is 120 cm tall.

In addition to the constant involvement of users, context of use analyses were performed. 
Regarding the physical environment, the height and range of interaction points were considered. Also, standard dimensions in the laboratory (walkway widths, door widths, etc.), the visual appearance and special features (e.g., typical colors, symbols), the volume and acoustic background, and the lighting conditions (day/ night/ special lighting conditions e. g. green light for certain assays) were taken into account. Organizational points that could be identified are, e.g., typical core working hours or shift work, public areas and non-public areas. Different user groups that may encounter and interact with the robot were identified (primary users (trained), secondary users (non-trained personnel: cleaning staff, visitors, etc.)). 

The eight potential users, who interacted directly with the robot, helped to decide the positioning of several, physical features. In a qualitative, iterative user survey they were asked to position features on the physical design (construction) of the robot (e.g., tablet location, orientation, and access, button positioning). Participants agreed on the following:
\begin{enumerate}
    \item the direct interaction points should not be positioned within the direct reach of the robot's manipulator,
    \item the controls should be positioned in a comfortable position for humans,
    \item the controls should always be accessible and visible.
\end{enumerate}

\begin{figure}[h]
\centering
\includegraphics[height=200pt]{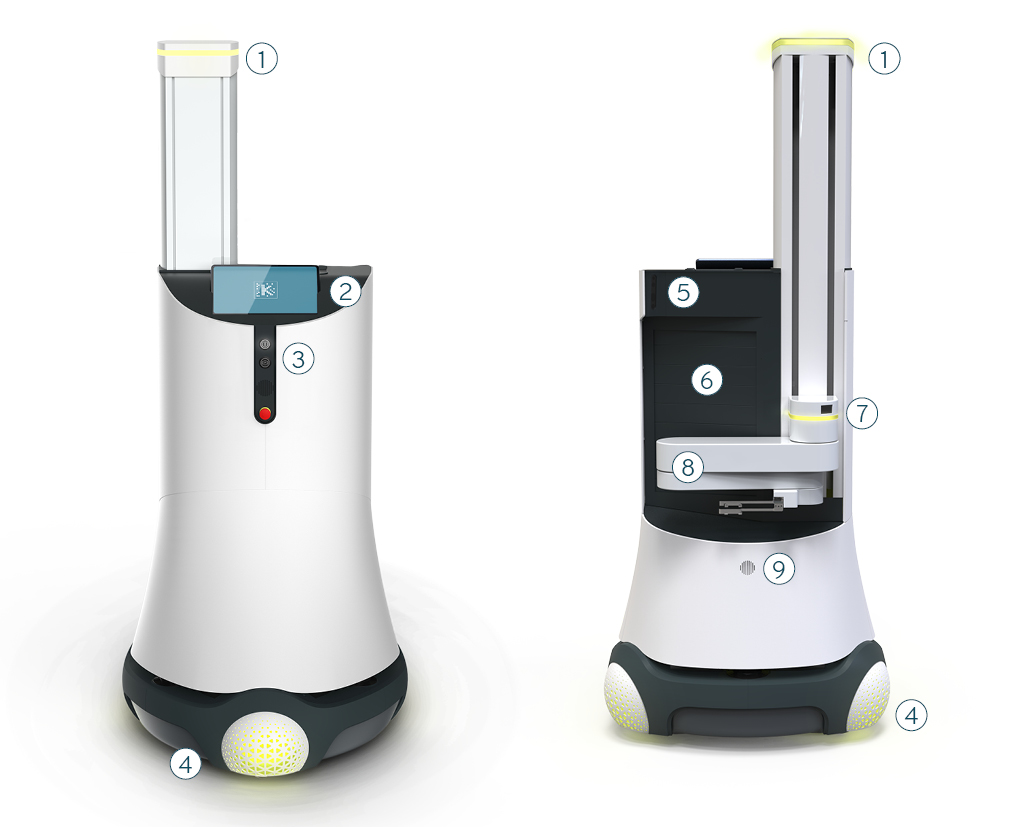}
\caption{Kevin 2.0´s Hardware Parts, left, Kevin seen from the back, right, from the front: 1 Headlight; 2 Detachable Tablet; 3 Buttons, Microphone and Speaker; 4 Omnidirectional Base with Lights and Sensors; 5 Camera and light; 6 Lockable Plate Hotel; 7 Camera; 8 Four-Axis SCARA Robot Sample Handler; 9 Speaker}
\label{fig:hardwareParts}
\end{figure}

In result, the manipulator was positioned \textit{in front} and features of direct human-robot interaction \textit{in the back} cf.~Figure \ref{fig:hardwareParts}. 

In order to support the counterpart in the human-robot interaction with orientation and interaction, visual cues are integrated into the design, which are intended to facilitate the affordance. For example, the robot's front is understood as the side that faces forward when driving. In the case of moving creatures and objects, the shape, positioning of features and the representation of the axes of motion enable conclusions to be drawn about subsequent movements.  In the design of the housing, great attention was paid to the representation of symmetry, the distinction between front and back, and the visualization of the axis of motion.

Speakers were positioned at the \textit{front} and at the \textit{back} of the robot, allowing sound to be perceived from all directions. The speakers are centrally located on Kevin, on the one hand for practical reasons and on the other hand to ensure that Kevin does not appear too large and therefore intimidating due to the sound origin. 

The structure and axes of motion of the manipulator are clearly visible. From the user tests, it was identified that a light element should be placed on the manipulator, which lights up when the manipulator is moving. In addition, a second light element was installed on top of the tower of the manipulator, which is at eye-level, visible from all sides, and visible from a distance. The three bottom light elements are part of the base and also visible from all sides, cf.~Fig.~\ref{fig:hardwareParts}.

\section{PRELIMINARY STUDY} 
\label{preStudy}

By the end of 2020 a preliminary study investigated the design of the auditory and visual output modalities. Based on the gathered information through interviews and workshops, three modality designs with different levels of complexity were created. These designs were compared through a within-subject user testing following a Wizard-of-oz study design. Five participants (four male and one female) were part of the study, all of them belonging to the Fraunhofer Institute, working in a laboratory environment. Each of them received the same two tasks, one to be executed in seated position and a second task, in which they had to move around the room.

\begin{figure}[h]
\centering
\includegraphics[width=0.48\textwidth]{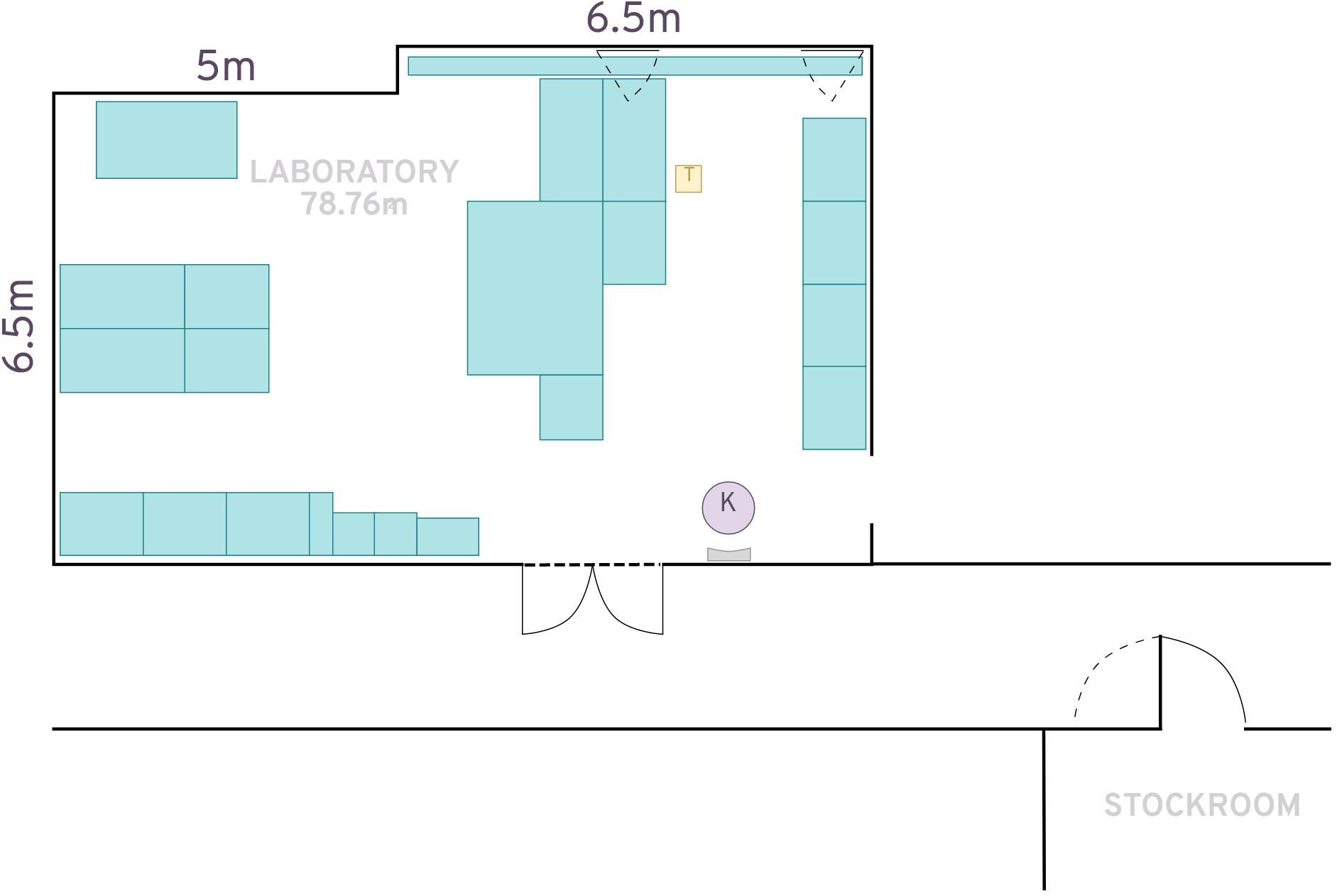}
\caption{Ground plan of the laboratory.}
\label{fig:groundplan}
\end{figure}

The user test took place in a life science laboratory of the Fraunhofer Institute, cf. Figure \ref{fig:groundplan}. The ventilation system and an extractor bonnet were switched on to create a realistic surrounding. The sound level of the laboratory was about 65 dB.

The procedure was divided into the following two parts:
\begin{enumerate}
    \item[A)] Firstly the user was given a task where he/she needed to concentrate on. During this task, the robot was driving around them.
    \item[B)] In the second task, the user needed to get something across the corridor and had to move in the same area as the robot.
\end{enumerate}   
Thereby, the aim was to recreate two key aspects of a normal working day in the laboratory.

\subsubsection{Three design variants}

To develop a recommendation for the design of the modalities of the laboratory robot Kevin, a minimal communication design without targeted communication was tested against two other designs, labelled "medium communication" and "maximum communication". In an interview the participants should display their personal preference for each state that has been communicated.

\begin{figure}
     \centering
     \begin{subfigure}[b]{0.3\columnwidth}
         \centering
         \includegraphics[width=.95\linewidth]{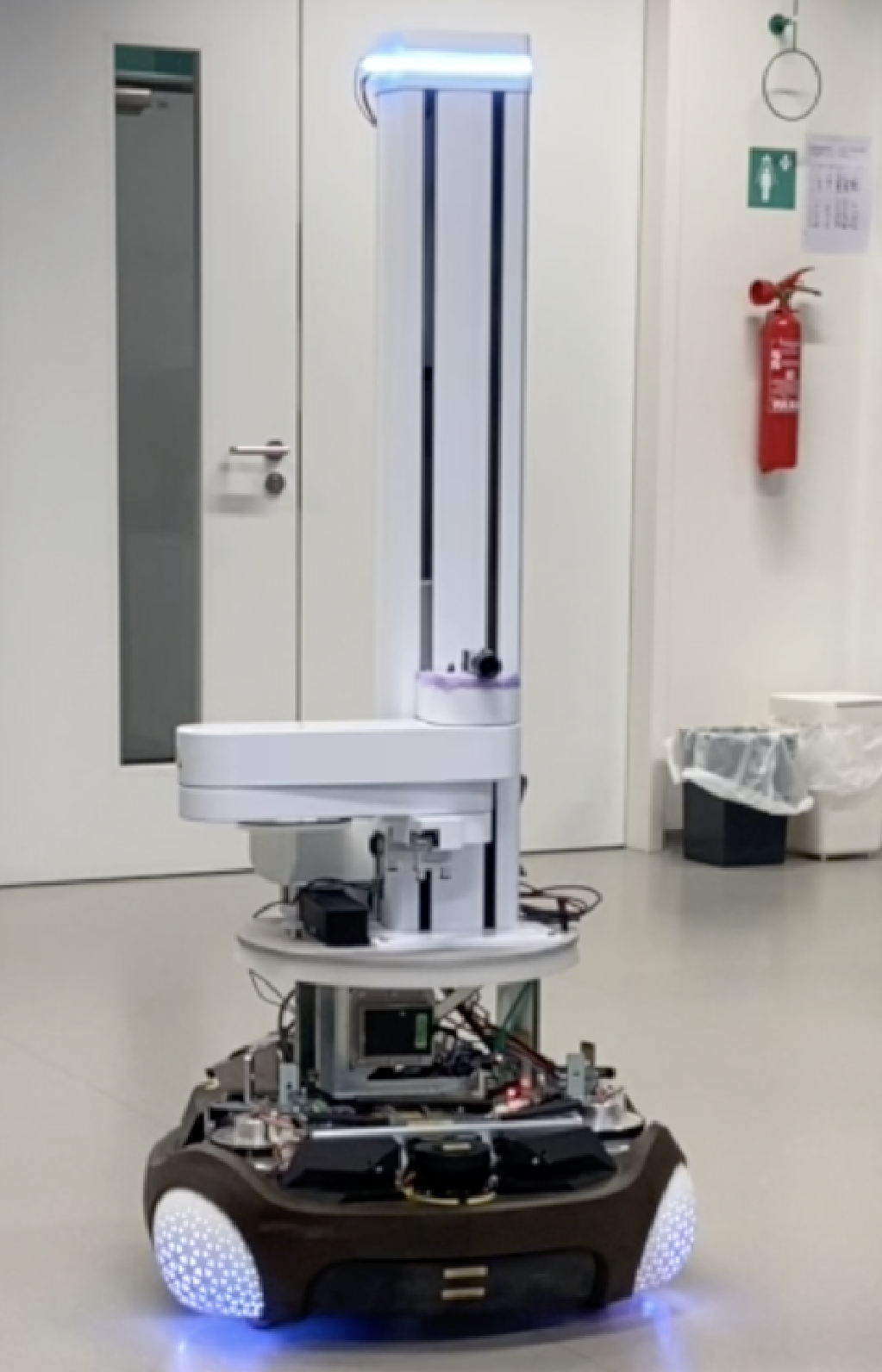}
         \caption{\small Minimal com. design: White light representing all states.}
         \label{fig:minimalDesign}
     \end{subfigure}
     \hfill
     \begin{subfigure}[b]{0.3\columnwidth}
         \centering
         \includegraphics[width=.95\linewidth]{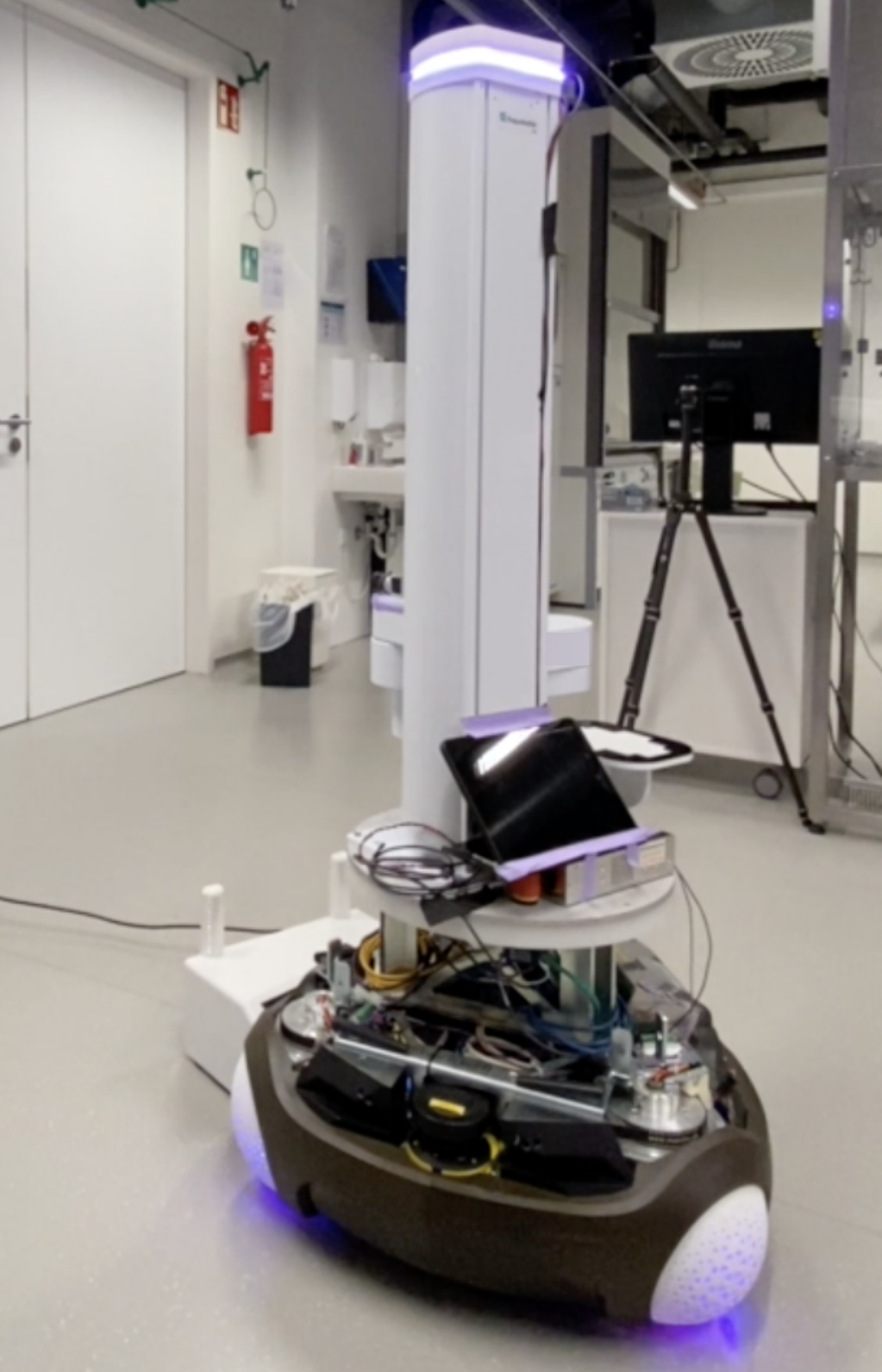}
         \caption{\small Medium com. design: Purple light representing the charging state.}
         \label{fig:mediumDesign}
     \end{subfigure}
     \hfill
     \begin{subfigure}[b]{0.3\columnwidth}
         \centering
         \includegraphics[width=.95\linewidth]{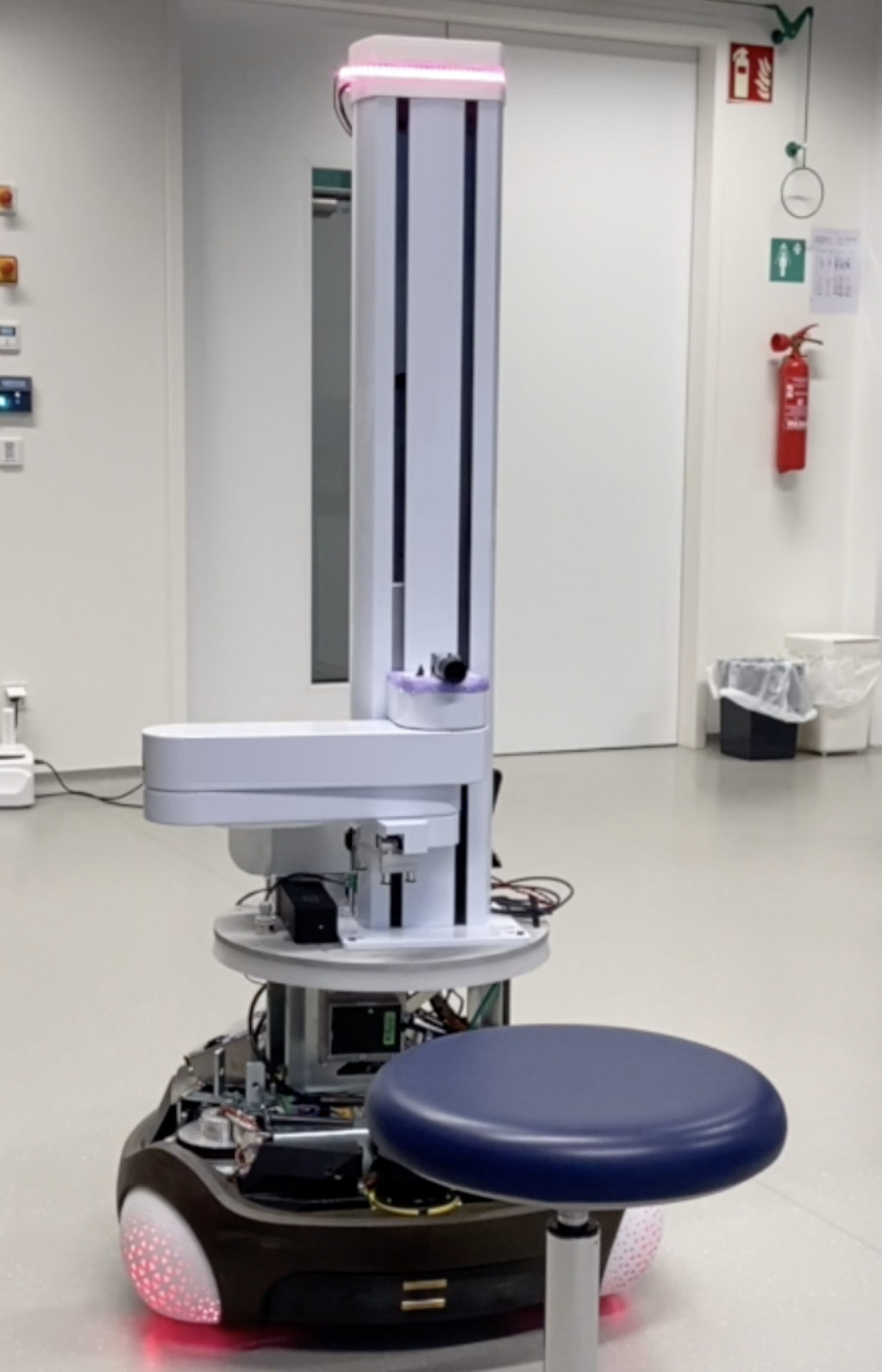}
         \caption{\small Maximum com. design: Red flashing light representing the charging state.}
         \label{fig:maximumDesign}
     \end{subfigure}
        \caption{The three different communication designs used within the preliminary study.}
        \label{fig:threeDesigns}
\end{figure}

\begin{itemize}
    \item The \textbf{minimal communication} level with white light only, cf. Figure \ref{fig:minimalDesign}.
    \item The \textbf{medium communication} level with a combination of
static light, cf. Figure \ref{fig:mediumDesign} and the display showing only
the error code.
    \item The \textbf{maximum communication} level consisted of light
including light sequences, non-verbal sounds and the
display containing additional information in all states
except for active and shut-down, cf. Figure \ref{fig:maximumDesign}.
\end{itemize}

Only the "maximum communication" design features so-called "non-verbal sounds", which consisted of a humming sound for driving and sounds similar to those that are used for feedback of a Bluetooth speaker, when buttons are pressed. 

\subsubsection{Findings}

Contrary to our expectation, continuously signaling the internal states of the robot seems unnecessary in most circumstances and a limited communication design fits the robot's appearance and situational context better. Only occasionally, the users required additional information that should be presented in a concise and intuitive manner. Accordingly, the medium communication design variant was preferred by most participants over the minimal and maximum design variants.

In summary, it was helpful to involve the potential users in the design process of a mobile robot to make it more acceptable and comprehensible. To avoid confusion and support the users understanding of the robot in the best way possible, a targeted design of the modalities is strongly recommended.

\section{CONCLUSIONS}

This paper reported on the two-step design process of the mobile laboratory robot Kevin. During the research and its development, the user-centered process was important for us to technically adapt the robot to the conditions of the laboratory. Kevin is able to support the user by taking over non-value adding tasks. Analysing the context of use and the target group and aligning the robot accordingly has been an important part of the process.
The key finding of the preliminary study is the importance of the targeted design regarding the individual modalities avoiding an information overload for the laboratory workers.

A limiting factor of our preliminary study is the low number of participants and their occupational background. Clearly, with only five participants the findings are hardly statistically significant.

As a next step, we intend to test the final design in a realistic setting. In the same vain, it will be important to conduct a long-term evaluation of the application of Kevin in a life science laboratory. This aims to ensure that the design will not be obstructive to the user in everyday working life.


\section*{APPENDIX}



\bibliographystyle{plain}
\bibliography{References}

\begin{thebibliography}{10}

\bibitem{baraka2018mobile}
Kim Baraka and Manuela~M Veloso.
\newblock Mobile service robot state revealing through expressive lights:
  formalism, design, and evaluation.
\newblock {\em International Journal of Social Robotics}, 10(1):65--92, 2018.

\bibitem{becker2011studying}
Christian Becker-Asano, Takayuki Kanda, Carlos Ishi, and Hiroshi Ishiguro.
\newblock Studying laughter in combination with two humanoid robots.
\newblock {\em AI \& society}, 26(3):291--300, 2011.

\bibitem{Roboticarm2011}
R.~Bingel-Erlenmeyer, V.~Olieric, J.~P.~A. Grimshaw, J.~Gabadinho, X.~Wang,
  S.~G. Ebner, A.~Isenegger, R.~Schneider, J.~Schneider, W.~Glettig,
  C.~Pradervand, E.~H. Panepucci, T.~Tomizaki, M.~Wang, and C.~Schulze-Briese.
\newblock Sls crystallization platform at beamline x06da—a fully automated
  pipeline enabling in situ x-ray diffraction screening.
\newblock {\em Crystal Growth \& Design}, 11(4):916--923, 2011.

\bibitem{Blow}
Mike Blow, Kerstin Dautenhahn, Andrew Appleby, Chrystopher~L. Nehaniv, and
  David Lee.
\newblock The art of designing robot faces: Dimensions for human-robot
  interaction.
\newblock In {\em Proceedings of the 1st ACM SIGCHI/SIGART Conference on
  Human-Robot Interaction}, HRI '06, page 331–332, New York, NY, USA, 2006.
  Association for Computing Machinery.

\bibitem{nonverbalSignals_2016}
Elizabeth Cha and Maja Mataric.
\newblock Using nonverbal signals to request help during human-robot
  collaboration.
\newblock pages 5070--5076, 10 2016.

\bibitem{repetitiveTask_2016}
S.~Cremer, L.~Mastromoro, and D.~O. Popa.
\newblock On the performance of the baxter research robot.
\newblock pages 106--111, 2016.

\bibitem{dautenhahn_2013}
Kerstin Dautenhahn.
\newblock {\em Human-robot interaction.}
\newblock The Interaction Design Foundation.

\bibitem{dautenhahn2002design}
Kerstin Dautenhahn.
\newblock Design spaces and niche spaces of believable social robots.
\newblock In {\em Proceedings. 11th ieee international workshop on robot and
  human interactive communication}, pages 192--197. IEEE, 2002.

\bibitem{fernandez2018passive}
Rolando Fernandez, Nathan John, Sean Kirmani, Justin Hart, Jivko Sinapov, and
  Peter Stone.
\newblock Passive demonstrations of light-based robot signals for improved
  human interpretability.
\newblock In {\em 2018 27th IEEE International Symposium on Robot and Human
  Interactive Communication (RO-MAN)}, pages 234--239. IEEE, 2018.

\bibitem{fong2003survey}
Terrence Fong, Illah Nourbakhsh, and Kerstin Dautenhahn.
\newblock A survey of socially interactive robots.
\newblock {\em Robotics and autonomous systems}, 42(3-4):143--166, 2003.

\bibitem{Groom}
Victoria Groom, Leila Takayama, Paloma Ochi, and Clifford Nass.
\newblock I am my robot: The impact of robot-building and robot form on
  operators.
\newblock In {\em 2009 4th ACM/IEEE International Conference on Human-Robot
  Interaction (HRI)}, pages 31--36, 2009.

\bibitem{haring2011creation}
Markus H{\"a}ring, Nikolaus Bee, and Elisabeth Andr{\'e}.
\newblock Creation and evaluation of emotion expression with body movement,
  sound and eye color for humanoid robots.
\newblock In {\em 2011 RO-MAN}, pages 204--209. IEEE, 2011.

\bibitem{Herberg}
Jonathan~S. Herberg, Dev~C. Behera, and Martin Saerbeck.
\newblock Eliciting ideal tutor trait perception in robots pinpointing
  effective robot design space elements for smooth tutor interactions.
\newblock In {\em 2013 8th ACM/IEEE International Conference on Human-Robot
  Interaction (HRI)}, pages 137--138, 2013.

\bibitem{SBSDimensions}
American Nationals~Standards Institutes.
\newblock Ansi/slas 1-2004: Microplates — footprint dimensions.
\newblock Technical report, Washington, D.C., 2004.

\bibitem{ishiguro2001robovie}
Hiroshi Ishiguro, Tetsuo Ono, Michita Imai, Takeshi Maeda, Takayuki Kanda, and
  Ryohei Nakatsu.
\newblock Robovie: an interactive humanoid robot.
\newblock {\em Industrial robot: An international journal}, 2001.

\bibitem{kittmann2015let}
Ralf Kittmann, Tim Fr{\"o}hlich, Johannes Sch{\"a}fer, Ulrich Reiser, Florian
  Wei{\ss}hardt, and Andreas Haug.
\newblock Let me introduce myself: I am care-o-bot 4, a gentleman robot.
\newblock {\em Mensch und computer 2015--proceedings}, 2015.

\bibitem{Lange2013}
Wolfgang Lange and Armin Bundesanstalt für Arbeitsschutz und~Arbeitsmedizin
  Windel.
\newblock {\em Kleine ergonomische Datensammlung}.
\newblock TÜV-Media, Köln, 15., aktualisierte aufl. edition, 2013.

\bibitem{aRobotCompanion}
Patrizia Marti and Leonardo Giusti.
\newblock A robot companion for inclusive games: A user-centred design
  perspective.
\newblock In {\em 2010 IEEE International Conference on Robotics and
  Automation}, pages 4348--4353, 2010.

\bibitem{norman2013}
Donald~A. Norman.
\newblock {\em The design of everyday things}.
\newblock Basic Books, [New York], 2013.

\bibitem{mci/Pollmann2019}
Kathrin Pollmann, Nora Fronemann, Nektaria Tagalidou, and Daniel Ziegler.
\newblock Bedürfnisbasierte personalisierung für die soziale mensch-roboter
  interaktion.
\newblock In {\em Mensch und Computer 2019 - Workshopband}, Bonn, 2019.
  Gesellschaft für Informatik e.V.

\bibitem{Rae2013}
Irene Rae and Leila Takayama.
\newblock The influence of height in robot-mediated communication.
\newblock pages 1--8, 03 2013.

\bibitem{SchwenkArras2014}
Markus Schwenk and Kai~O. Arras.
\newblock R2-d2 reloaded: A flexible sound synthesis system for sonic
  human-robot interaction design.
\newblock In {\em The 23rd IEEE International Symposium on Robot and Human
  Interactive Communication}, pages 161--167, 2014.

\bibitem{adaptBehavior_2019}
Gilbert Tang, Phil Webb, and John Thrower.
\newblock The development and evaluation of robot light skin: A novel robot
  signalling system to improve communication in industrial human–robot
  collaboration.
\newblock 56:85--94, 2019.

\bibitem{yilmazyildiz2016review}
Selma Yilmazyildiz, Robin Read, Tony Belpeame, and Werner Verhelst.
\newblock Review of semantic-free utterances in social human--robot
  interaction.
\newblock {\em International Journal of Human-Computer Interaction},
  32(1):63--85, 2016.

\end{thebibliography}
\end{document}